\pdfoutput=1
\documentclass{article}

\usepackage[final,nonatbib]{neurips_2020}

\usepackage[utf8]{inputenc} 
\usepackage[T1]{fontenc}    
\usepackage{hyperref}       
\hypersetup{
  pdftitle = {Open-domain Topic Identification of Out-of-domain Utterances using Wikipedia}
}
\usepackage{url}            
\usepackage{booktabs}       
\usepackage{amsfonts}       
\usepackage{nicefrac}       
\usepackage{microtype}      

\usepackage{graphicx}       
\usepackage{mathtools}      
\usepackage{array}

\DeclareMathOperator*{\argmin}{arg\,min}

\title{Open-domain Topic Identification of Out-of-domain Utterances using Wikipedia}

\author{
  Alexandry Augustin\thanks{Work done while working at Toshiba Research Europe.} \\
  University of Southampton\\
  Southampton, UK\\
  \texttt{aa7e14@soton.ac.uk}\\
  \And
  Alexandros Papangelis\footnotemark[1]\\
  Amazon Alexa AI\\
  Sunnyvale, USA\\
  \texttt{papangea@amazon.com}\\
  \And
  Margarita Kotti\\
  Toshiba Research Europe\\
  Cambridge, UK\\
  \texttt{margarita.kotti@crl.toshiba.co.uk}\\
  \AND
  Pavlos Vougiouklis\thanks{Work done while working at University of Southampton.}\\
  Huawei Technologies\\
  Edinburgh, UK\\
  \texttt{pavlos.vougiouklis@huawei.com}\\
  \And
  Jonathon Hare\\
  University of Southampton\\
  Southampton, UK\\
  \texttt{jsh2@ecs.soton.ac.uk}\\
  \And
  Norbert Braunschweiler\\
  Toshiba Europe Limited\\
  Cambridge, UK\\
  \texttt{norbert.braunschweiler@crl.toshiba.co.uk}\\
}

\begin{document}

\maketitle

\begin{abstract}
Users of spoken dialogue systems (SDS) expect high quality interactions
across a wide range of diverse topics. However, the implementation
of SDS capable of responding to every conceivable user utterance
in an informative way is a challenging problem. Multi-domain SDS
must necessarily identify and deal with out-of-domain (OOD) utterances
to generate appropriate responses as users do not always know in
advance what domains the SDS can handle. To address this problem,
we extend the current state-of-the-art in multi-domain SDS by estimating
the topic of OOD utterances using external knowledge representation
from Wikipedia. Experimental results on real human-to-human
dialogues showed that our approach does not degrade domain prediction performance when compared to the base model. But more significantly, our joint training achieves more accurate predictions of the nearest Wikipedia article by up to about 30\% when compared to the benchmarks. 
\end{abstract}

\section{Introduction\label{sec:Introduction}}
Human-to-human dialogues are often composed of multiple sub-dialogues
bridging a wide range of \emph{topics}. In contrast, multi-domain
spoken dialogue systems (SDS) are often designed to operate over
a limited and static set of predefined topics called \emph{domains} (e.g. hotel
or restaurant booking) to improve on performance \cite{lane_out--domain_2007}.
The limited domain coverage in multi-domain SDS has proven to be
a challenge for inexperienced users as they do not necessarily know
in advance what domains the SDS is able to handle efficiently. Such
users may attempt to formulate utterances that cannot be handled by
the SDS. These are referred to as out-of-domain (OOD) utterances.
To illustrate the ubiquity of OOD, around 13\% of utterances in the
TourSG dataset
used in the fourth edition of the Dialogue State Tracking Challenge\footnote{\url{http://www.colips.org/workshop/dstc4/}}
(DSTC4) were OOD. For this reason, graceful handling of OOD utterances
is crucial to provide robustness to SDS against unexpected user
inputs while providing helpful responses.

A number of fallback strategies have been employed to generate responses
to OOD utterances. They range from very simplistic answers such
as ``I do not understand. Could you please rephrase?'' \cite{ultes_pydial:_2017},
to more sophisticated chatbots that generate responses to hold the
users' attention \cite{papaioannou_combining_2017,charras_comparing_2018}.
These approaches however do not always provide informative answers
to the users. Some efforts have
been devoted towards integrating large knowledge graphs to allow
users more freedom in expressing their intents \cite{papangelis_ld-sds:_2017}.
Other efforts have focused on leveraging large amount of unstructured
data to extract answers (e.g. sentences or phrases), or to synthesise
new answers \cite{lowe_incorporating_2015}. In all of these
scenarios, the identification of open-domain topics can be leveraged
to provide informative responses to OOD utterances \cite{khatri_contextual_2018}.
For example, access to the knowledge graph can be restricted only
to areas relevant to the open-domain topic \cite{waltinger_interfacing_2011}.
Similarly, keywords or keyphrases can be extracted from unstructured
documents and used as answers based on the open-domain topic \cite{frank_domain-speci_1999}.
Finally, answers can be synthesised from entire or sections of documents
which are topically relevant \cite{mckeown_towards_1999,krishna_generating_2018}.

Now, a growing body of research explores the use of topics to enable informative responses to OOD utterances in SDS \cite{lagus_topic_2002,bang2015open,waltinger_interfacing_2011,kawahara_topic_2004,lane_out--domain_2007,komatani_managing_2008}. 
A simple way to identify such topics is to use approaches
based on word \cite{ultes_pydial:_2017} or phrase \cite{boros_linguistic_1999}
matching. Alternatively, probabilistic topic modelling
\cite{blei_probabilistic_2012} offers another powerful, albeit more complex, strategy to identify
open-domain topics \cite{guo_topic-based_2018,khatri_contextual_2018}. These methods however
do not usually take advantage of the time-dependence between utterances in the dialogue history to improve on topic identification.
Since utterances are often short (e.g. \textquotedblleft thank you\textquotedblright{}
or ``yes''), one should account for long-term dependencies between utterances to provide context. Furthermore, topic models provide no guarantee that the topics inferred will be interpretable to humans \cite{chang_reading_2009}.
This is particularly true when faced with short utterances as large amounts of data is required for these models to be accurate. Approaches based on recurrent architectures \cite{hochreiter_long_1997} have been proposed \cite{kim_exploring_2016} 
to account for the time-dependency between utterances. In particular Kim et al. \cite{kim_exploring_2016}
used an LRCN-based approach \cite{7558228,karpathy_deep_2015} to classify the domain of each utterance accounting for the dialogue history up to that particular
utterance. The Kim et al. \cite{kim_exploring_2016} method however only classify the domain and do not reveal new information on OOD utterances.

Against this background, we argue that a key research direction
to enable informative responses to OOD utterances in SDS is the joint
identification and tracking of both the domain and topic of each
utterance to ensure robustness and accuracy in a wide range of dialogue
settings \cite{jokinen_adaptive_2002}. In particular, when utterances
fall into domains that the SDS can handle, they are dealt with efficiently.
On the other hand, when utterances are OOD, an informative answer
is still provided based on the topic identified. As such, we extend
Kim et al. \cite{kim_exploring_2016} to support
open-domain topic tracking. We fit the model to real world human-to-human dialogues which
have been manually labelled with the domain of each utterance and
automatically labelled with the most relevant topic. Our method is based on the assumption
that external encyclopedic knowledge from Wikipedia can be used to
identify relevant topics for any given utterance. This assumption
is common and has been employed in a number of prior related works
\cite{vaart_asymptotic_2000,schonhofen_identifying_2006,banerjee_clustering_2007,breuing_lets_2012,bhatia_automatic_2016}.
Since Wikipedia is constructed and maintained collaboratively by a
large number of volunteers, it provides huge amounts of encyclopedic
knowledge. This enables the construction of a shared semantic
space in which new and unseen utterances are mapped to the closest
article in terms of semantic similarity.

In more detail, we make the following contributions to the state-of-art:
(i) we define a new model that jointly learns, for the first time,
both the domain and topic of each utterance, (ii) we empirically show
that our approach generates comparable performance when identifying
the domains when compared to the Kim et al. \cite{kim_exploring_2016} model, and achieves up to about 30\% improvement in accuracy against the benchmarks
when predicting the topics, (iii) we show that the exclusion of previous utterances when training leads to suboptimal performance.

The remainder of this paper is organised as follows. We first introduce
the Kim et al. \cite{kim_exploring_2016} model in Section \ref{sec:Preliminaries}.
We then detail our extension in Section \ref{sec:model}. In Section
\ref{sec:Evaluation}, we present the results of our experimental
evaluation. We conclude in Section \ref{sec:conclusion}.

\section{Preliminaries\label{sec:Preliminaries}}

\begin{figure*}[t]
\begin{center}
\includegraphics[scale=0.5]{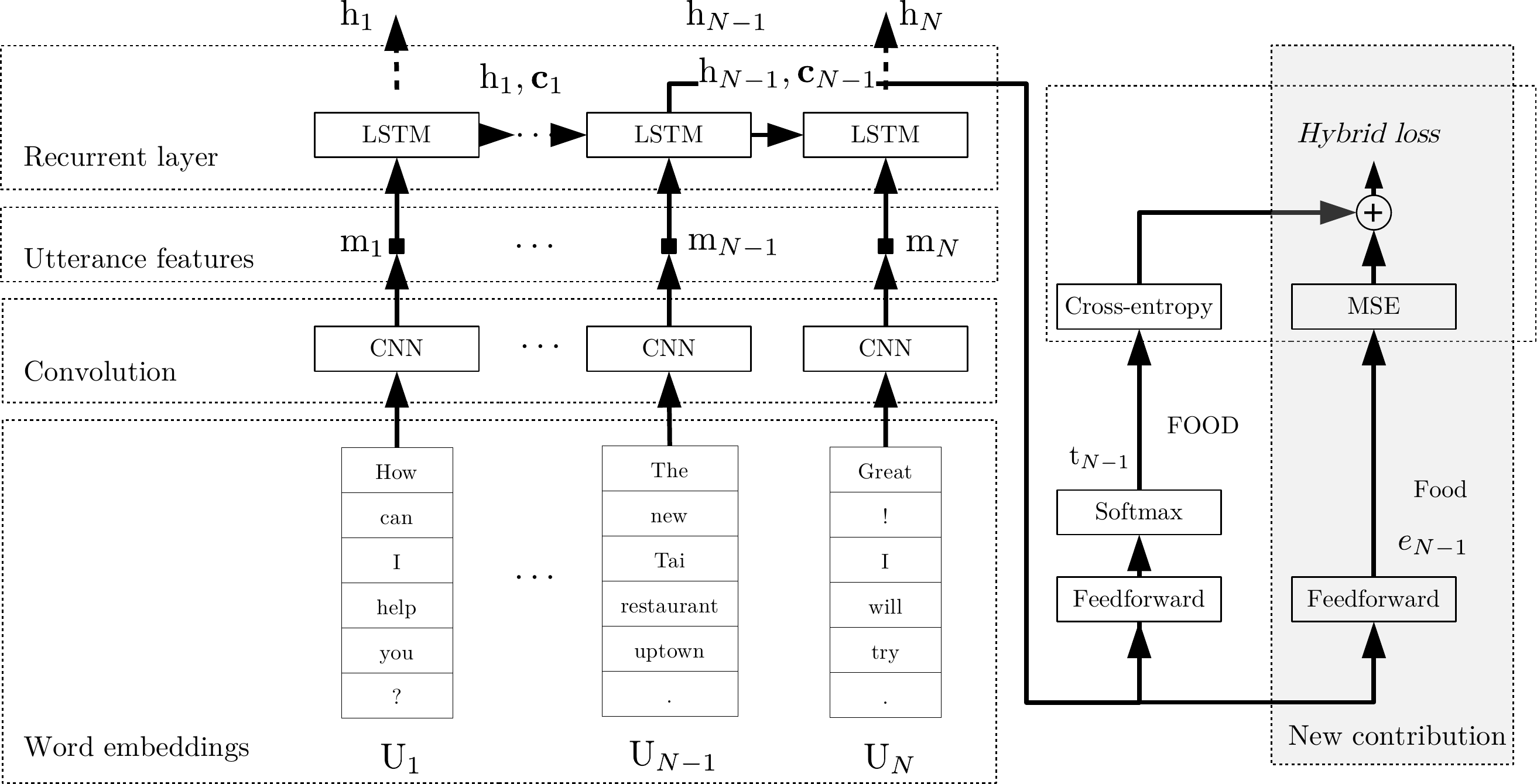}
\end{center}
\protect\caption{\label{fig:model}LRCN architecture combining a CNN and an RNN for
utterance classification and regression. During training, the model
takes as input a set of utterances and outputs a domain and a topic
embedding.}
\end{figure*}

The Kim et al. \cite{kim_exploring_2016} model takes as input an utterance
and performs a classification of its domain accounting for the dialogue
history up to that particular utterance. In more detail, the architecture
is based on a long recurrent convolutional neural network (LRCN)
\cite{7558228,karpathy_deep_2015,vougiouklis2016neural}, that
is, a composition of a CNN \cite{LeCun:1999:ORG:646469.691875}
and an LSTM \cite{hochreiter_long_1997} (Figure \ref{fig:model}).
While CNN computes a fixed-size feature vector for each utterance,
the LSTM captures the dependency in time between the feature vectors.
More precisely, let $\mathcal{D}=\left\{ \left(\mathbf{u}_{t},\mathbf{x}_{t}\right):0\leq t\leq\left|\mathcal{D}\right|\right\} $
be a dataset of utterance and domain pairs\@. The model takes as
input an utterance $\mathbf{u}_{t}\in\mathbb{N}^{N}$ at time $t$
where $N$ is the number of words in the utterance. Each element of $\mathbf{u}_{t}$
contains the index of a word in the vocabulary. An embedding layer
\cite{mikolov_efficient_2013} then map each discrete index to a
continuous embedding vector resulting in a matrix $\mathbf{U}_{t}\in\mathbb{R}^{N\times K}$
where $K$ is the embedding size. The embedding layer is shared across
timesteps such that a given word contributes changes during training
to the same embedding regardless of the timestep at which it appears.
Furthermore, the word embeddings are either initialised randomly and
learned during training, or fine-tuned from pre-trained word embeddings
\cite{mikolov_efficient_2013} to speed up the training \cite{kim_exploring_2016}.
The CNN used is based on the architecture proposed by Kim et al. \cite{kim_convolutional_2014}
and takes as input the utterance $\mathbf{U}_{t}$. It performs a
convolution over $\mathbf{U}_{t}$ by sliding a set of filters of
given height $M$ and fixed width $K$ (the same as width as the input)
over the rows of $\mathbf{U}_{t}$ . Each filter is applied to $\mathbf{U}_{t}$
to generate a feature map $\mathbf{\mathbf{v}}_{t}$. Each feature
element $\mathbf{v}_{t,i}$ of a feature map is generated from a subregion
$\mathbf{U}_{t,i:i+M-1}=\left\{ U_{t,k}:\mbox{for }i\leq k\leq i+M-1\right\} $
of the input utterance from the $i$-th to the $\left(i+M-1\right)$-th
row such that 
\[
\mathbf{v}_{t,i}=f\left(\mathbf{W}\mathbf{U}_{t,i:i+M-1}+\mathbf{b}\right)
\]
 where $f$ an activation function (e.g. ReLU or sigmoid). The weights
$\mathbf{W}\in\mathbb{R}^{M\times K}$ and biases $\mathbf{b}\in\mathbb{R}^{M}$
of each filter are shared for all $i$. To capture the salient
feature of each feature map, a global max-pooling is applied resulting
in a scalar $m_{t}=\max\left(\mathbf{v}_{t}\right)$for each filter.
Since utterances at different time steps are likely to be of different
lengths, global max-pooling is particularly suited as it guarantees
that the output of the CNN will always stay the same regardless of
the input size. The resulting scalar from the global max-pooling
of each filter are then concatenated into a single feature vector
$\mathbf{m}_{t}$. In turn, the utterance feature vector $\mathbf{m}_{t}$
is presented as input to an LSTM which captures the dependency
in time between the utterances via a hidden state vector $\mathbf{h}_{t}\in\mathbb{R}^{H}$
and a cell state vector $\mathbf{c}_{t}\in\mathbb{R}^{H}$. 
The hidden state $\mathbf{h}_{t}$ of the LSTM is then presented
to a dropout layer \cite{hinton_improving_2012} to improve on training
performance and generalisation. Finally, a feedforward layer is added
to compute the domain classification scores over the $D$ domains.
These scores are then fed into a softmax layer that produces the predictive
distributions over the domains such that
\[
\mathbf{x}_{t}=\mbox{softmax}\left(\mathbf{W}_{xh}\mathbf{h}_{t}+\mathbf{b}_{x}\right)
\]
with weights $\mathbf{W}_{xh}\in\mathbb{R}^{D\times H}$ and bias
$\mathbf{b}_{x}\in\mathbb{R}^{D}$. The output of the model at
time $t$ is the domain distribution $\mathbf{x}_{t}$ associated
with the current utterance $\mathbf{u}_{t}$. The model is trained
using stochastic gradient descent in a supervised learning fashion,
using a cross-entropy loss function between the ground truth domain
$\hat{\mathbf{x}_{t}}$ (i.e. a one-hot vector) and the predicted
domain distribution $\mathbf{x}_{t}$ at each time-step $t$
\begin{equation}
H_{\theta_{x}}=-\sum_{i}\mathbf{x}_{t,i}\mbox{log}\hat{\mathbf{x}}_{t,i} \label{eq:tracker-entropy}
\end{equation}
where $\theta_{x}$ is the set of all weights and biases prior to
$\mathbf{x}_{t}$. The benefit of this combined architecture is its
ability to be trained end-to-end. That is, the CNN learns the input
features that are relevant for the sequence labelling. However, such
a model doesn't address open-domain topic tracking as is, due to
the inherent limitation of classifying utterances into domains that
have to be known a priori. 

\section{Joint Tracking\label{sec:model}}

Our proposed model extends the one presented in the previous section
to deal with settings where utterances are assigned to both domains
and topics. To do so, in addition to learning the domain of each
utterance, we also learn a continuous mapping of similarity between
utterances and Wikipedia articles\footnote{\url{en.wikipedia.org}}
under a common semantic space. Specifically, our method consists of
two main steps. In the first step, we automatically build a training
set of utterance and Wikipedia article pairs. This is done offline
prior to training our model. In the second step, we extend the structure
of Kim et al. \cite{kim_exploring_2016} to interpolate the mapping identified
in the previous step to new and unseen utterances, taking into account
the dialogue history up to that utterance.

More specifically, in the first step, we rank the most relevant Wikipedia
articles to each training utterance. To perform the ranking, we use
the term frequency inverse document frequency (TF-IDF) algorithm \cite{manning_introduction_2008}.
TF-IDF is a prevailing technique in information retrieval and suited
to our setting for its simplicity given the large amount of Wikipedia
articles considered. Words in an utterance with a high TF-IDF
score imply a strong relationship with the Wikipedia article they
appear in. Furthermore, independently of the ranking, we compute
a document embedding for each Wikipedia article using the doc2vec
algorithm \cite{le_distributed_2014}. Since doc2vec is applicable
to texts of any length (although longer semantic units yield more
accurate vectors), it can readily be used to compute the embeddings
of the Wikipedia articles in our setting (hereafter referred to as
\emph{topic embeddings}). Let $d_{i}$ be the topic embedding of
the $i$-th article in the Wikipedia dataset $\mathcal{D}_{W}$ computed
from the doc2vec algorithm. The objective of doc2vec is to minimise
the cross-entropy loss when predicting the missing word $w_{i,j}$
for all words $j$ in each article $i\in\mathcal{D}_{W}$, that is
minimising 
\[
-\mbox{log}p\left(w_{i,j}|\cdots,w_{i,j-1},w_{i,j+1},\cdots,d_{i}\right).
\]
 The topic embeddings $d_{i}$ are computed once, separately of
our model. Now that we have both a ranking of Wikipedia articles
per utterance, and a topic embedding for each article, we can associate
a training target for each utterance in the training set. Each training
target for an utterance consists of the topic embedding of the top
matching article or the average top-$k$ topic embeddings.

\begin{figure}[t]
\begin{centering}
\includegraphics[scale=0.35]{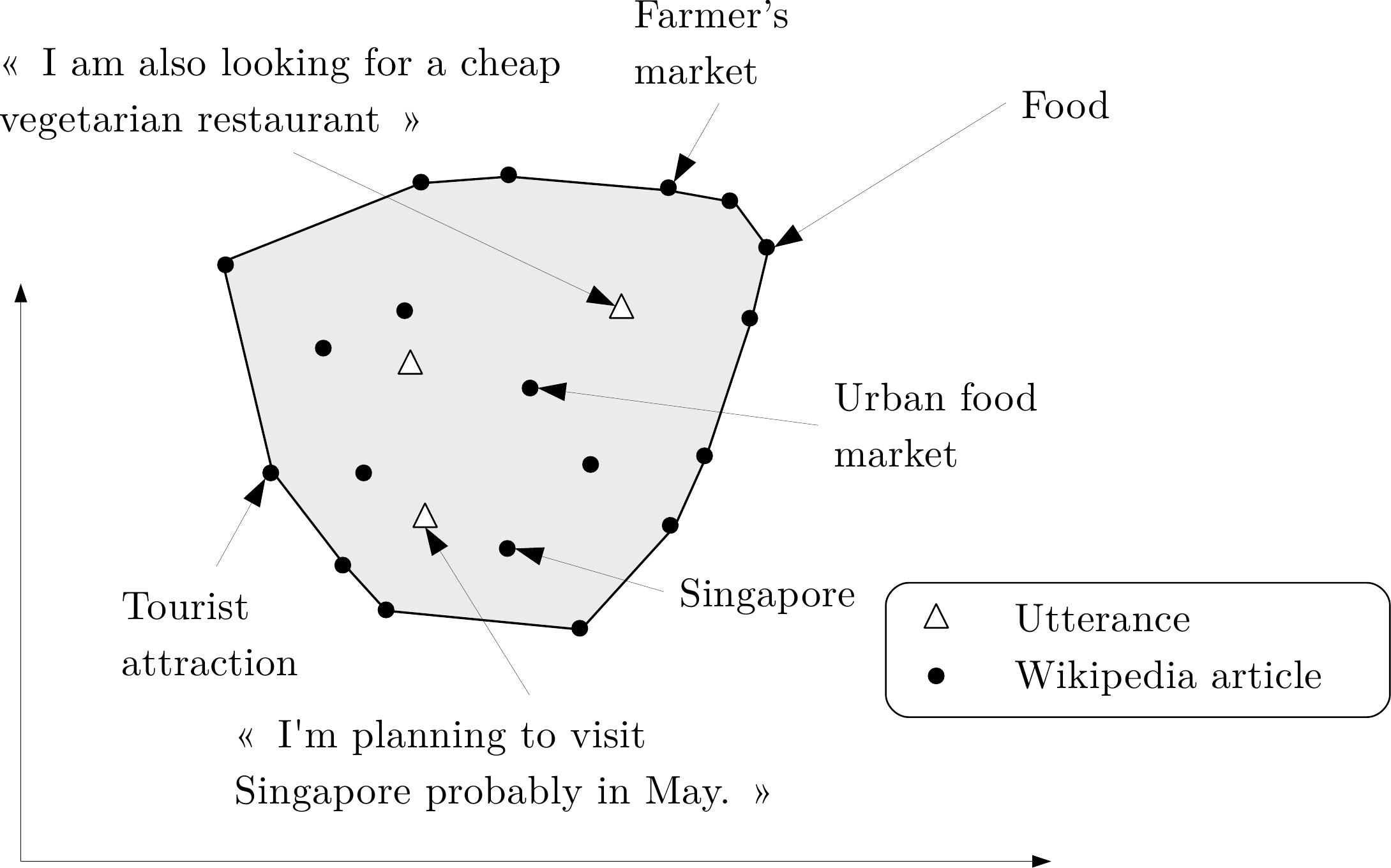}
\par\end{centering}
\protect\caption{\label{fig:open-domain-semantic-space}Illustration of a two-dimensional
semantic space with associated convex Hull (shaded area) of the topic embeddings.}
\end{figure}

In the second step, we extend the Kim et al. \cite{kim_exploring_2016} model
to account for open-domain topics by performing a regression on the known target topic embeddings.
This is achieved by using a fully connected feedforward layer 
linked to the output of the LSTM
\[
\mathbf{y}_{t}=\mathbf{W}_{yh}\mathbf{h}_{t}+\mathbf{b}_{y}
\]
with weights $\mathbf{W}_{yh}\in\mathbb{R}^{K\times H}$ and bias
$\mathbf{b}_{y}\in\mathbb{R}^{K}$. Its role is to learn a mapping
between the output of the LSTM and the target topic embeddings. In
particular, the model learns to embed each utterance into the semantic
space consisting of the topic embeddings constructed by the doc2vec
algorithm (Figure \ref{fig:open-domain-semantic-space}). In such
a semantic space, a continuous similarity measure (e.g. Euclidean
distance or cosine similarity) is used to compute the distance between
each utterance embedding and the closest topic embedding. When performing inference on unseen utterances, the solution
lies within the convex Hull formed by topic embeddings in the training
set (Figure \ref{fig:open-domain-semantic-space}). Given the set
of all topic embeddings $S=\left\{ d_{i}:\mbox{for all \ensuremath{i} in \ensuremath{\left|\mathcal{D}_{W}\right|}}\right\} $,
the convex Hull is the intersection of all convex sets in $S$
\[
C=\left\{ \sum_{j}\lambda_{j}d_{j}:\lambda_{j}\geq0\mbox{ for all \ensuremath{j} and }\sum_{j}\lambda_{j}=1\right\} .
\]
Therefore, the extent of the training data defines the solution
space for the topic embeddings.

To train the model, we first concatenate all the dialogue sessions
with each other, and then slide a context window of length $H$ across
a fixed number of utterances. In other word, the input of the model
at each timestep $t$ consists of the set of utterances $U_{t-H:t}$
for $t\in\left\{ 1,\cdots,\left|\mathcal{D}\right|\right\}$. This
prevents the model's complexity (i.e. its number of parameters) being
dependent on the shape of the dataset, and in particular to the maximum
length of the sessions which may be large. We assess the loss using
a squared error objective function between the predicted topic embedding
$\mathbf{y}_{t}$ and the ground truth topic target $\hat{\mathbf{y}}_{t}$
\begin{align}
\mbox{SE}_{\theta_{y}}=||\mathbf{y}_{t}-\hat{\mathbf{y}}_{t}||^{2}\label{eq:tracker-mse}
\end{align}where $\theta_{y}$ is the set of all weights and biases prior to $\mathbf{y}_{t}$. The model is jointly optimised end-to-end using a multi-objective learning function, encompassing errors not only from the topic classification, but also errors from topic regression
\begin{align}
\argmin_{\theta_{x},\theta_{y}}\lambda_{x}H_{\theta_{x}}+\lambda_{y}\mbox{SE}_{\theta_{y}}
\end{align}where $H_{\theta_{x}}$ is given by Equation \ref{eq:tracker-entropy},
$\mbox{SE}_{\theta_{y}}$ by Equation \ref{eq:tracker-mse}, and the
weights $\lambda_{x},\lambda_{y}\in\mathbb{R}$.

\section{Experimental Evaluation\label{sec:Evaluation}}

In this section we report our performance results on both domain
and topic identification.

\subsection{Dataset\label{sub:datasets}}

Our experiments use a total of two datasets.

\textbf{TourSG.} This corpus (released as part of the DSTC4\footnote{\url{http://www.colips.org/workshop/dstc4/}}
competition) is composed of 35 manually transcribed dialogue sessions
between tour guides and tourists in Singapore. Each of the 31,034
utterances has been annotated with one of nine domains: ATTRACTION
(39.2\%), TRANSPORTATION (13\%), OTHER (12.7\%), FOOD (12.4\%),
ACCOMMODATION (11.3\%), SHOPPING (5.7\%), ITINERARY (2.3\%), CLOSING
(1.7\%) and OPENING (1.6\%). The dataset has a vocabulary size
of 6,035 words. The average length of an utterance is $9.25\pm8.01$
words, and the average length of a session is $887\pm185$ utterances.

\textbf{Wikipedia.} The pre-processing of our training data requires
the Wikipedia dataset\footnote{\url{https://en.wikipedia.org/wiki/Main_Page}}.
The dataset is composed of 4.5 million articles of the English
Wikipedia, and has a vocabulary size of 2 million words. No other
information from the dataset was used other than the title and the
raw text content of each article.

\subsection{Performance Metrics\label{sub:preformance-metrics}}

The performance in identifying both domains and topics
are assessed using standard multi-class classification measures. In
particular, to identify the topics, we assess the models at recovering
the Wikipedia articles identified by TF-IDF. That is, for each utterance
in the test set we first compute its embeddings, and then we use the
Euclidean distance to identify the nearest neighbouring topic embeddings
from the Wikipedia articles. We label the learned embeddings as correctly
classified if the nearest topic embedding matches with the one identified
by TF-IDF, otherwise they are labelled as an incorrect classification.

\subsection{Benchmarks\label{sub:Benchmarks}}

We compare performance against each component of our proposed architecture.
The strength of each benchmark lies in the potential inclusion of
the dialogue history and the mechanism by which the utterance features
are calculated.

\textbf{CNN.} This benchmark does not take into account the dialogue
history. It first computes the utterance features, and then performs
classification and/or regression using a feedforward layer followed
by a softmax layer for the classification, and a feedforward layer
alone for the regression.

\textbf{LSTM. }This benchmark takes into account temporal dependencies
between utterances. However, instead of taking as input the utterance
features computed from the CNN, it uses pre-trained embeddings from
the doc2vec algorithm.

\begin{table*}[t]
\tiny
\centering  
\begin{tabular}{>{\raggedright}p{3cm}lll>{\raggedright}p{3cm}>{\raggedright}p{3cm}}
\toprule 
 &  & \multicolumn{2}{c}{Domain} & \multicolumn{2}{c}{Topic}\tabularnewline
Utterance  & Speaker  & Actual  & Predicted  & Actual & Predicted \tabularnewline
\midrule 
Hi, good morning-  & Guide  & OPENING  & OPENING  & Hi convoys  & Hi convoys \tabularnewline
\#uh good afternoon.  & Guide  & OPENING  & OPENING  & Afternoon (disambiguation)  & Phyllomacromia aureozona \tabularnewline
This is \#uh tour guide one.  & Guide  & OPENING  & OPENING  & List of events at the Jacksonville Coliseum  & List of events at the Jacksonville Coliseum \tabularnewline
Lynnette here.  & Guide  & OPENING  & OPENING  & Violin Concerto (Bernard Tan)  & Udpura \tabularnewline
Yah.  & Tourist  & OPENING  & FOOD  & South Sea Tales (London collection)  & South Sea Tales (London collection) \tabularnewline
Hi Lynnette this is participant number eleven.  & Tourist  & OPENING  & OPENING  & Hi convoys  & Bukit Nanas Monorail station \tabularnewline
And can I have your name please?  & Guide  & OPENING  & OPENING  & Please  & Phyllomacromia aureozona \tabularnewline
Yah.  & Tourist  & OPENING  & SHOPPING  & South Sea Tales (London collection)  & South Sea Tales (London collection) \tabularnewline
Yah, this is participant number eleven.  & Tourist  & OPENING  & ITINERARY  & South Sea Tales (London collection)  & List of flag bearers for Singapore at the Olym... \tabularnewline
Okay, and how can I help you?  & Guide  & OPENING  & OPENING  & Okay (disambiguation)  & Okay (disambiguation) \tabularnewline
Yah I'm planning to have \#um- to go around Asi...  & Tourist  & ITINERARY  & ITINERARY  & South Sea Tales (London collection)  & Uruguay (disambiguation) \tabularnewline
And what attractions can I go in Singapore?  & Tourist  & ATTRACTION  & ATTRACTION  & Outline of Singapore  & Outline of Singapore \tabularnewline
Okay.  & Guide  & ATTRACTION  & ATTRACTION  & Okay (disambiguation)  & Okay (disambiguation) \tabularnewline
\#Uh which part of the year are you planning to...  & Guide  & ITINERARY  & FOOD  & Planning cultures  & European countries by electricity consumption ... \tabularnewline
\#Um maybe this \#um last week of May.  & Tourist  & ITINERARY  & ITINERARY  & Maybe  & Stranger Things (disambiguation) \tabularnewline
This year.  & Tourist  & ITINERARY  & OPENING  & List of communes in Puerto Rico  & 1904 Philadelphia Phillies season \tabularnewline
Okay.  & Guide  & ITINERARY  & ATTRACTION  & Okay (disambiguation)  & Okay (disambiguation) \tabularnewline
\bottomrule
\end{tabular}
\protect\caption{\label{tab:dialogue-test-set-1}Predictions from the jointly trained
LRCN on a randomly selected dialogue session in the test set of the
TourSG dataset.}
\end{table*}

\textbf{Random.} This benchmark assigns to each utterance a domain
and a Wikipedia article at random with equal probabilities. 

Each benchmark is trained in three different configurations: domain
classification only (D), topic regression only (T), and both\footnote{Excluding the random benchmark.}
(D+T).

\subsection{Experimental Setting\label{sub:experimental-setting}}

To ensure generalisation and avoid a dependence of our model’s
parameters to the structure of the training dataset, we concatenate the
35 dialogues sessions of the TourSG dataset into a single contiguous
sequence of 31,034 utterances. This enables us to use a context window
of fixed size irrespective of the number and length of each dialogue session.
For this reason, our approach is applicable to any dialogue dataset, provided
a topic and a topic class are assigned to each utterance. It is worth noting that by concatenating the TourSG dataset, the context window will intermittently overlap with opening and closing utterances from adjacent dialogue sessions at training time. Although this consequence represents a small fraction of the training data, it may impact prediction performance of the opening and closing utterances at test time. One approach would be to pad each dialogue sessions by a small but fixed amount (depending of the context window's size) of missing values (i.e. NULL) prior to concatenation. For example, a context window of 10 utterances would require a padding of 9 missing values between each dialogue session.

As such, we unroll the LSTM for 20 timesteps which is a reasonable range
for recurrent architectures to perform well \cite{bahdanau_neural_2014}. We train the models in a supervised setting and divide our collection of 31,034
contiguous utterances into training (60\%, i.e. 18,620
utterances), validation (20\%, i.e. 6,207 utterances), and test sets
(20\%, i.e. 6,207 utterances). We further set the hidden state size of the LSTM
to 300, the embedding size to 200, and the batch size to 5 utterances. We use 64
filters of height 1 and stride 1, with global max-pooling for the CNN. The CNN
and LRCN are initialised with pre-trained
GloVe\footnote{\url{https://nlp.stanford.edu/projects/glove/}} word embeddings for
faster convergence. We use the implementation of the doc2vec algorithm provided by the
Gensim\footnote{\label{gensim}\url{https://radimrehurek.com/gensim/}} library to compute the pre-trained
topic embeddings and utterance embeddings for the LSTM benchmark. Furthermore, we
make use of PV-DM variant of the doc2vec algorithm as it has been shown to consistently
perform better than PV-DBOW \cite{le_distributed_2014}. We exclude Wikipedia articles
of less than 50 words due to the limitations of doc2vec with short-length documents
\cite{de_boom_learning_2015}. We use the Gensim's implementation of the TF-IDF algorithm to map the utterances to the Wikipedia articles. After this mapping, we are left with
4,409 unique Wikipedia articles out of the 4.5 million initial candidates. We use the
topic embedding of the top matching Wikipedia article ($k=1$) as training target for each utterance.
Finally, we set the dropout probability to 80\%, and elect the Adam optimiser
\cite{kingma_adam:_2014-1} with a learning rate to 0.001.

\subsection{Results\label{sub:results}}

\begin{table*}[t]
\centering
\begin{tabular}{lcccccccc}
\toprule 
 & \multicolumn{4}{l}{Domains} & \multicolumn{4}{l}{Topics}\tabularnewline
Models  & A  & F  & P  & R  & A  & F  & P  & R \tabularnewline
\midrule 
Random (D) & 11.11 & 8.89 & 10.97 & 10.81 & - & - & - & -\tabularnewline
CNN (D)  & 50.17  & 47.61  & 56.48  & 50.17  & -  & -  & -  & - \tabularnewline
LSTM (D)  & 35.28  & 34.31  & 34.89  & 35.28  & -  & -  & -  & - \tabularnewline
LRCN (D)  & \textbf{51.44}  & \textbf{50.78}  & 52.25  & \textbf{51.44}  & -  & -  & -  & - \tabularnewline
\midrule 
Random (T) & - & - & - & - & 0 & 0 & 0 & 0\tabularnewline
CNN (T)  & -  & -  & -  & -  & 17.82  & 16.33  & 15.77  & 17.82 \tabularnewline
LSTM(T)  & -  & -  & -  & -  & 13.86  & 13.18  & 15.33  & 13.86 \tabularnewline
LRCN (T)  & -  & -  & -  & -  & 24.75  & 24.50  & 24.88  & 24.75 \tabularnewline
\midrule 
CNN (D+T)  & 51.09  & 48.26  & \textbf{57.43}  & 51.09  & 18.81  & 17.19  & 16.04  & 18.81 \tabularnewline
LSTM(D+T)  & 38.51  & 36.52  & 37.16  & 38.51  & 19.80  & 18.38  & 18.05  & 19.80 \tabularnewline
LRCN (D+T)  & 50.28  & 49.49  & 50.71  & 50.28  & \textbf{30.69 } & \textbf{30.24 } & \textbf{31.55 } & \textbf{30.69 }\tabularnewline
\bottomrule
\end{tabular}
\caption{\label{tab:tracker-accuracy-results}Performance comparison between
our approach (i.e. LRCN (D+T)) and the benchmarks. The letter(s) in
parentheses indicate if training has been performed on domain classification
only (D), topic regression only (T), or both (D+T). The letters A,
F, P, R stands for accuracy, F1-score, precision and recall respectively.
The best performing models are highlighted in bold.}
\end{table*}

\begin{table}[t]
\centering
\begin{tabular}{>{\raggedright}p{1cm}>{\raggedright}p{2cm}llll}
\toprule 
Domain  & Model  & A  & F  & P  & R \tabularnewline
\midrule 
ACC  & LRCN (D)  & \textbf{44.73}  & \textbf{61.81}  & 100.00  & \textbf{44.73} \tabularnewline
 & LRCN (D+T)  & 38.97  & 56.08  & 100.00  & 38.97 \tabularnewline
ATTR & LRCN (D)  & \textbf{69.71}  & \textbf{82.15} & 100.00  & \textbf{69.71} \tabularnewline
 & LRCN (D+T)  & 69.37  & 81.92  & 100.00  & 69.37 \tabularnewline
CLOSE  & LRCN (D)  & \textbf{50.52}  & \textbf{67.12}  & 100.00  & \textbf{50.52} \tabularnewline
 & LRCN (D+T)  & 47.42  & 64.34  & 100.00  & 47.42 \tabularnewline
FOOD  & LRCN (D)  & \textbf{46.01}  & \textbf{63.02}  & 100.00  & \textbf{46.01} \tabularnewline
 & LRCN (D+T)  & 44.60  & 61.69  & 100.00  & 44.60 \tabularnewline
ITI  & LRCN (D)  & \textbf{38.14}  & \textbf{55.21}  & 100.00  & \textbf{38.14} \tabularnewline
 & LRCN (D+T)  & 28.81  & 44.74  & 100.00  & 28.81 \tabularnewline
OPEN  & LRCN (D)  & \textbf{75.61}  & \textbf{86.11}  & 100.00  & \textbf{75.61} \tabularnewline
 & LRCN (D+T)  & 69.51  & 82.01  & 100.00  & 69.51 \tabularnewline
OTHER  & LRCN (D)  & \textbf{26.03}  & \textbf{41.31}  & 100.00  & \textbf{26.03} \tabularnewline
 & LRCN (D+T)  & 25.39  & 40.49  & 100.00  & 25.39 \tabularnewline
SHOP  & LRCN (D)  & \textbf{35.49}  & \textbf{52.39}  & 100.00  & \textbf{35.49} \tabularnewline
 & LRCN (D+T)  & 32.76  & 49.36  & 100.00  & 32.76 \tabularnewline
TRSP  & LRCN (D)  & 44.59  & 61.68  & 100.00  & 44.59 \tabularnewline
 & LRCN (D+T)  & \textbf{45.98}  & \textbf{63.00}  & 100.00  & \textbf{45.98} \tabularnewline
\bottomrule
\end{tabular}
\caption{\label{tab:classification-accuracies-breakdown}Classification performance
breakdown per domain of the LRCN when trained jointly (D+T), and on domains only (D). The best performing settings are
highlighted in bold.}
\end{table}

Table \ref{tab:tracker-accuracy-results} compares the performance
of our model and the benchmarks when trained in the three different
configurations: domain classification only, topic regression only,
and both. We first observe that all models outperform the random
benchmark by up to 20\%. In particular, we observe that the challenge
of predicting the correct Wikipedia article is so significant that
the random benchmark is unable to recover any articles at all, achieving
an accuracy, recall and precision of zero. Such a task is in effect
equivalent to performing a multi-class classification with about four
thousand alternatives (i.e. the number of unique Wikipedia articles
in the training set) compared to the more manageable nine alternatives
in the domain prediction problem. Furthermore, we note that the
combined LRCN approach clearly outperforms each of its components in
isolation across all training configurations. This confirms that
the CNN is indeed learning useful utterance features, that in turn
sees their dependencies in time successfully captured by the LSTM.

Table \ref{tab:classification-accuracies-breakdown} shows the performance breakdown per domain of the LRCN when jointly trained, and trained on domains only. Table \ref{tab:dialogue-test-set-1} depicts an example of our model's predictions (i.e. ``Predicted domain'' and ``Predicted topic'') for a randomly selected dialogue session in the test set. The ``Actual domain'' field refers to the annotated labels from the TourSG dataset. While ``Actual topics'' refers to the Wikipedia article identified by TF-IDF.

\section{Conclusion\label{sec:conclusion}}

We introduced a novel architecture that for the first time simultaneously tracks both the domain and the topic of each utterance in a dialogue session. Our key premises were that: (i) the handling of these two settings is essential to achieve both efficiency and accuracy in the response generated by downstream systems, and (ii) the title and content of Wikipedia articles are a reasonable proxy for the topic of an utterance. We showed experimentally on a real-world dataset that our approach of jointly training the LRCN generates comparable performance when identifying the domain than the components in isolation, but more significantly, it is up to about 30\% more accurate when predicting the nearest Wikipedia article. As a future step, it would be beneficial to add a measure of uncertainty in the topic predictions to lessen the impact of prediction errors in downstream systems. Addressing this issue is challenging however as the large number of parameters in our approach prevents the direct use of Bayesian inference techniques \cite{mackay_practical_1992,pearce_uncertainty_2018}.

\bibliography{main}
\bibliographystyle{abbrv}

\end{document}